\documentclass{article} 
\usepackage{nips13submit_e,times}
\usepackage{hyperref}
\usepackage{url}
\usepackage{graphicx}
\usepackage{caption}
\usepackage{subcaption}
\usepackage{multirow}

\title{On Binary Classification with Single--Layer Convolutional Neural Networks}

\author{
Soroush Mehri\\
University of Montreal, Canada\\
\texttt{mehris@iro.umontreal.ca} \\
}

\nipsfinalcopy 

\begin{document}

\maketitle
\begin{abstract}
Convolutional neural networks are becoming standard tools for solving object recognition and visual tasks. However, most of the design and implementation of these complex models are based on trail-and-error. In this report, the main focus is to consider some of the important factors in designing convolutional networks to perform better. Specifically, classification with wide single--layer networks with large kernels as a general framework is considered. Particularly, we will show that pre--training using unsupervised schemes is vital, reasonable regularization is beneficial and applying of strong regularizers like dropout could be devastating. Pool size is also could be as important as learning procedure itself. In addition, it has been presented that using such a simple and relatively fast model for classifying cats and dogs, performance is close to state-of-the-art achievable by a combination of SVM models on color and texture features.
\end{abstract}

\section{Introduction}
Convolutional neural networks, first proposed by [1], have achieved state-of-the-art performance in many tasks (including [2] and [3]) by introducing local connectivity, parameter sharing, and pooling/subsampling hidden units in compare to fully--connected networks. However, there is little knowledge on how these complex models work which makes the design and development of these models based on experience.

Current solution to high-level tasks such as classification in computer vision and object recognition is to use features obtained by stacking multiple convolutional layers with higher layers capturing increasingly more complex representation [4]. Design of each layer depends on different choices including but not limited to kernel size and stride, pooling shape and stride, in addition to different regularization methods that could play a essential role in the end result and generalization to unseen input.

Jarret et al. [5] studied different pooling strategies, along with different form of rectification and normalization between layers of features, and Boureau et al. [6]-[7] studied different coding schemes ans pooling strategies. In this work, the main focus is on the effect of these decisions on shallow but wide convolutional networks for binary classification. Although this is in essence similar and closely related to analysis of single--layer networks in unsupervised feature learning by Coates et al. [8], here we present analysis for convolutional networks for object recognition, particularly with bigger kernel size than usual recent trend, to compensate the loss of capturing long distance dependencies normally gained by stacking multiple levels of representation and  effect of pre--training in such structures using unsupervised feature extractors, namely zero-bias autoencoders introduced in 2014 by Memisevic et al. [9].

\section{Task and Dataset}
The challenge is to distinguish between cats and dogs on the Asirra CAPTCHA [10], a dataset for an already-finished competition hosted by Kaggle. The task is easy for humans, while designed to be presumably difficult to be accomplished automatically [10].

Before introducing as a Kaggle competition, 82.7\% was the highest accuracy by combination of support-vector machine classifiers trained on color and texture features of images [11], while according to the leaderboard being almost solved by 98.914\% accuracy [12] using OverFeat library [13].

The dataset used in following experiments are from the same dataset except that only the training set is available. The data consists of 25000 variable size, JPEG format, RGB images, with balanced classes. The data set is then divided into 20000, 2500, and 2500 respectively for train, validation, and test set. The images are 360 in height and 404 pixels in width on average.

\section{Training Framework}
In this section a common framework is explained to be explored throughout the experiments and create a baseline to compare the performance. For the most part, unless explicitly stated, the pipeline is as follows.
\subsection{Model Architecture}
Smallest dimension of each image is scaled to 200 and randomly cropped with 192 size. The architecture of the network is mostly one layer of valid convolutional network with 512 $16\times 16$, $24\times 24$, or $32\times 32$ kernels (to ''see'' more region of input), $8\times 8$ pooling shape and pooling stride, untied biases, followed by a 1024-dimension fully-connected layer and on top, a softmax layer to output the probability of corresponding classes. Rectified linear units [14] are used in all the layers.

Despite the computational inefficiency of small kernel strides, effectiveness of its value in performance is practically proved in [8]. Hence, this hyper--parameter is set to smallest value of one for all experiments.

\subsection{Learning Scheme}
All models are trained with early--stopping, learning rate of 0.001, 0.9 momentum, $L_2$ regularization (only for free parameters) or dropout only on fully-connected layer. Mini--batch size is set to 1000 to reduce the overhead of I/O and dealing with bottleneck of transferring (between hard disk, RAM, and GPU memory) and on-the-fly pre--processing.

At this point this is worthwhile noting that in spite of having simpler features in compare to deep architectures, this one benefits from much lower inference time by the means of utilizing the parallel computation of hardware, here a GTX TITAN Black graphic card, and not suffering from vanishing gradient expected in deep models.

\subsection{Unsupervised Feature Extraction and pre--training}
The bigger kernel, the more parameters to learn from the same amount of data and learning large kernels without pre--training cannot give any boost in performance unless trained on lots of data, for a long time.

\begin{figure}[htbp]
\centering
\begin{subfigure}[b]{0.31\textwidth}
        \includegraphics[width=1.7in]{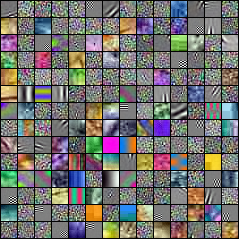}
        \caption{Some of 16x16 kernels}
        \label{zae_16}
    \end{subfigure}
    \begin{subfigure}[b]{0.31\textwidth}
        \includegraphics[width=1.7in]{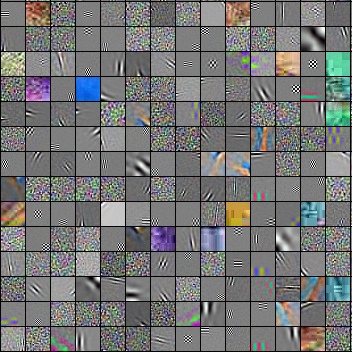}
        \caption{Some of 24x24 kernels}
        \label{zae_24}
    \end{subfigure}
    \begin{subfigure}[b]{0.31\textwidth}
        \includegraphics[width=1.7in]{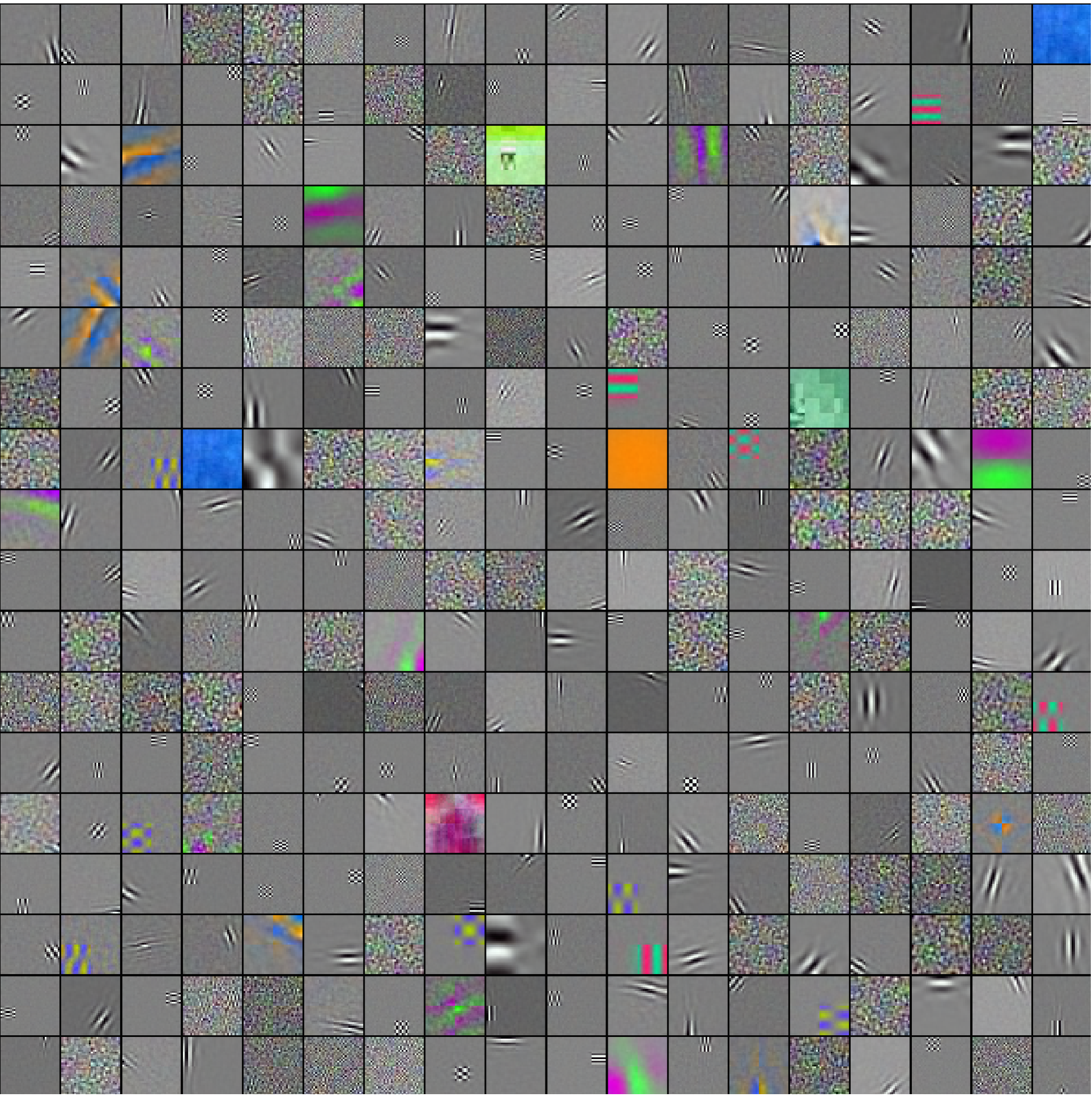}
        \caption{Some of 32x32 kernels}
        \label{zae_32}
    \end{subfigure}
\caption{Extracted features from ZAE. Clearly, capturing structures in the natural images, even JPEG compression artifacts in the from of local checkerboard patterns with sharp edges. A wide range of other filters are present, including local high-frequency patterns, color gradients, and edge detectors. (Best viewed in color; high quality images are available in \href{http://www-etud.iro.umontreal.ca/~mehris/16.png}{http://www-etud.iro.umontreal.ca/$\sim$mehris/\{16.png, }\href{http://www-etud.iro.umontreal.ca/~mehris/24.png}{24.png, }\href{http://www-etud.iro.umontreal.ca/~mehris/32.png}{32.png\}}).}
\label{zae_filters}
\end{figure}

Similar but not identical to [15], here the pre--training starts with randomly selecting one million patches from raw images with the size of kernel ($N=w\times w \times 3$), reducing dimensionality to 2000, 1250, 500 dimensions respectively for $w=32$, $w=24$, and $w=16$. Then trained a zero--bias autoencoder [9] for each with threshold of one and code layer of size 4096 (i.e. 8 times the number of kernels in CNN). Figure~\ref{zae_filters} shows some of the learned features for different kernel sizes to be used in CNN.

\section{Experiments and Analysis}
Experiments are divided in following branches, in each branch goal is to keep all the parameters and factors constant while changing one side of model. Implementation of following experiments have been done using Theano [16] and Pylearn2 [17] softwares.

\subsection{Effect of pre--training}
On one hand using bigger kernels results in bigger field of view, on the other hand they are much harder and more expensive to train as the number of parameters in the network grows, demanding larger dataset and more time to train. To exaggerate the effect of pre--training, experiments designed on biggest kernel size, $w=32$. Table~\ref{pretrain} clearly shows that networks with large kernel size can actually benefit from pre--training, when in this experiment there is approximately 25\% potential boost in moving from vanilla CNN to pre--trained one (compare exp1 and exp3). Remember that SVM classifiers achieved 17.3\% test error rate which is not far from 22.6\% achieved with single--layer CNN.

However, this brings us to the next point, whether learn the pre--trained parameters and randomly initialized parameters (here fully-connected layer and weights in softmax layer) jointly. Test error rate for models with ''frozen'' kernels (where during training gradient signals are just to adjust the other weights and biases) is much lower (compare exp2 and exp3).

Interestingly, this effect is also observable for a model with frozen initial random weights (exp14) in compare to letting the model dynamically go under fluctuations caused by strong stochastic approximate gradients (exp1). A natural next step is to do the fine--tuning of the whole network, allowing the different parts adjust to each other. In this experiment no gain is observed (exp3 and exp13).

Experiments 4 to 10 is then done based on the same way experiment 3 has been conducted, then the prediction of these methods once being treated as regression and averaged over different models, and once used in a voting system. It turned out that averaging (exp11) is more fruitful than voting over classifiers (exp12).

\begin{table}[t]
\caption{Effect of pre--training, fine--tuning and freezing kernels. All based on pipeline in section 3, 512 kernel size of $32\times 32$, $8 \times 8$ pool shape and stride, $L_2$ weight decay. (8 numbers appeared in parenthesis corresponds to 8 different model trained on 8 chunks of extracted features by ZAE)}
\label{pretrain}
\begin{center}
\begin{tabular}{|l|l|l|}
\hline
\textbf{\#} & \textbf{Model} & \textbf{Test error} \\ \hline
1 & No pre--training applied & 47.84\% \\ \hline
2 & With pre--training (1), No Freeze & 35.56\%\\ \hline
3 & With pre--training (1), Freeze & \textbf{22.60\%}\\
4 & With pre--training (2), Freeze &  24.60\%\\
5 & With pre--training (3), Freeze  &  23.30\%\\
6 & With pre--training (4), Freeze  &  23.56\%\\
7 & With pre--training (5), Freeze  &  24.32\%\\
8 & With pre--training (6), Freeze  &  25.16\%\\
9 & With pre--training (7), Freeze  &  23.80\%\\
10 & With pre--training (8), Freeze  &  24.36\%\\ \hline
11 & Averaging &  \textbf{21.56\%}\\
12 & Voting &  21.96\%\\ \hline
13 & Fine--tuning & \textbf{22.60\%} \\ \hline
14 & Random kernels, Freeze & 42.28\% \\ \hline
\end{tabular}
\end{center}
\end{table}

\subsection{Effect of Kernel Size}
The results in Table~\ref{kernel} are not consistent enough to draw a conclusion about the effect of kernel size. However, two opposing factors are crucial to consider. First, bigger kernels is harder to train, and might be oversimplifying the representation over a huge region of actual image, hurting the performance. Nonetheless, with same number of kernels, here 512, small kernels corresponding to bigger response maps, resulting in slower optimization over more parameters.
\begin{table}[t]
\caption{Effect of kernel size. All based on pipeline in section 3, 512 kernel size of $w\times w$, $8 \times 8$ pool shape and stride, $L_2$ weight decay, freezing kernels and no fine--tuning.}
\label{kernel}
\begin{center}
\begin{tabular}{|l|l|l|l|l|}
\hline
\textbf{\#} & \textbf{Kernel Size ($w$)} & \textbf{Test error} & \textbf{Validation error} & \textbf{Train error} \\ \hline
1 & 16 & 23.20\% & 25.08\% & 20.07\%\\
2 & 24 & 24.60\% & 26.00\% & 21.80\%\\
3 & 32 & \textbf{22.60\%} & 24.88\% & 21.61\%\\ \hline
\end{tabular}
\end{center}
\end{table}

\subsection{Effect of Regularization}
As a simple model with relatively simple features, this model needs the property of co--adaptation of units in fully--connected and softmax layer in order to capture more structured input. Hence, introducing this constraint through dropout[18] besides parameter sharing in CNN will cripple the model severely ($\sim$47\% test error rate). Practical results follow this analysis completely, resulting in not using dropout in any other experiments in this work. General form of regularization in this work is to use relatively high weight decay to close the gap between training and validation curves.

\subsection{Effect of Adding Extra Convolutional Layer and Pooling Size}
These set of experiments are a bit different in a sense that changes in pool shape is not the single difference, but another convolution layer (256 $4\times 4$ kernels with $2\times 2$ pooling shape and stride) between current one and fully--connected layer is added. From Table~\ref{poolsize}, not surprisingly it is clear that adding layer is helping the model (exp1 and exp4). From test errors it seems that the bigger the pool size is, the lower the error rate is. This holds but up to some point. When the pooling is getting closer to its extreme, the models looses more information than gaining performance by down--sampling.

By looking closely to the pre--trained $32\times 32$ kernels, $p=16$ looks reasonable. Most of the kernels, if are not dead, are local and not greater than half of kernel size, meaning when convolved on the input image with low stride, does not produce new values, and could be summarized in one value. Or kernels are mostly low frequency (e.g. color gradient), which makes the response map to change with respect to input really slowly. This method is not accurate, but could be used as a first hunch in hyper-parameter search.

One could argue that the model should get the same or more amount of information when using pooling size of less than 16, and would ignore the irrelevant information through learning. The point is that learning to ignore these information is much harder and time--consuming than not providing the network with it, making the learning time less.
\begin{table}[t]
\caption{Effect of pooling size. All based on pipeline in section 3, 512 kernel size of $32\times 32$, $p \times p$ pool shape and stride, $L_2$ weight decay.}
\label{poolsize}
\begin{center}
\begin{tabular}{|l|l|l|}
\hline
\textbf{\#} & \textbf{Model\/pool shape ($p$)} & \textbf{Test error} \\ \hline
1 & With pre--training (1), Freeze, $p=8$ & 22.60\%\\ \hline
2 & With added layer, $p=2$ & 28.00\%\\ 
3 & With added layer, $p=4$ & 23.08\%\\
4 & With added layer, $p=8$ & 20.12\%\\
5 & With added layer, $p=16$ & \textbf{19.32}\%\\ 
6 & With added layer, $p=32$ & 20.68\%\\ 
\hline
\end{tabular}
\end{center}
\end{table}

\section{Conclusion}
In this work, some of design choices for a wide single--layer convolutional neural network have been examined to potentially reduce the model selection time by picking more informed hyper--parameters. We studied the effect of pre--training, fine--tuning, kernel size, regularization and pool size. The best result with such a simple model is also comparable with the SVM results for the proposed dataset.

\subsubsection*{Acknowledgments}
The author would like to acknowledge the support of the following agencies for research funding and computing support: NSERC, Calcul Qu\'{e}bec, Compute Canada, the Canada Research Chairs and CIFAR. We would also like to thank the developers of Theano \footnote{http://deeplearning.net/software/theano/}, for developing such a powerful tool for scientific computing. Apart from that, I would like to thank Dr. Roland Memisevic for the rich course, sharing expertise, and valuable guidance on the subject at hand.

\subsubsection*{References}

\small{
[1] LeCun, Y., Boser, B., Denker, J. S., Henderson, D., Howard, R. E., Hubbard, W., \& Jackel, L. D. (1989). Backpropagation applied to handwritten zip code recognition. Neural computation, 1(4), 541-551.

[2] Krizhevsky, A., Sutskever, I., \& Hinton, G. E. (2012). Imagenet classification with deep convolutional neural networks. In Advances in neural information processing systems (pp. 1097-1105).

[3] Ciresan, D., Meier, U., \& Schmidhuber, J. (2012, June). Multi-column deep neural networks for image classification. In Computer Vision and Pattern Recognition (CVPR), 2012 IEEE Conference on (pp. 3642-3649). IEEE.

[4] Zeiler, M. D., \& Fergus, R. (2014). Visualizing and understanding convolutional networks. In Computer Vision–ECCV 2014 (pp. 818-833). Springer International Publishing.

[5] Jarrett, K., Kavukcuoglu, K., Ranzato, M., \& LeCun, Y. (2009, September). What is the best multi-stage architecture for object recognition?. In Computer Vision, 2009 IEEE 12th International Conference on (pp. 2146-2153). IEEE.

[6] Boureau, Y. L., Ponce, J., \& LeCun, Y. (2010). A theoretical analysis of feature pooling in visual recognition. In Proceedings of the 27th International Conference on Machine Learning (ICML-10) (pp. 111-118).

[7] Boureau, Y. L., Bach, F., LeCun, Y., \& Ponce, J. (2010, June). Learning mid-level features for recognition. In Computer Vision and Pattern Recognition (CVPR), 2010 IEEE Conference on (pp. 2559-2566). IEEE.

[8] Coates, A., Ng, A. Y., \& Lee, H. (2011). An analysis of single-layer networks in unsupervised feature learning. In International Conference on Artificial Intelligence and Statistics (pp. 215-223).

[9] Memisevic, R., Konda, K., \& Krueger, D. (2014). Zero-bias autoencoders and the benefits of co-adapting features. arXiv preprint arXiv:1402.3337.

[10] Elson, J., Douceur, J. R., Howell, J., \& Saul, J. (2007, October). Asirra: a CAPTCHA that exploits interest-aligned manual image categorization. In ACM Conference on Computer and Communications Security (pp. 366-374).

[11] Golle, P. (2008, October). Machine learning attacks against the Asirra CAPTCHA. In Proceedings of the 15th ACM conference on Computer and communications security (pp. 535-542). ACM.

[12] Dogs vs cats - private leaderbroad. Retrieved from https://www.kaggle.com/c/dogs-vs-cats/leaderboard

[13] Sermanet, P., Eigen, D., Zhang, X., Mathieu, M., Fergus, R., \& LeCun, Y. (2013). Overfeat: Integrated recognition, localization and detection using convolutional networks. arXiv preprint arXiv:1312.6229.

[14] Glorot, X., Bordes, A., \& Bengio, Y. (2011). Deep sparse rectifier networks. In Proceedings of the 14th International Conference on Artificial Intelligence and Statistics. JMLR W\&CP Volume (Vol. 15, pp. 315-323).

[15] Rifai, S., Bengio, Y., Courville, A., Vincent, P., \& Mirza, M. (2012). Disentangling factors of variation for facial expression recognition. In Computer Vision–ECCV 2012 (pp. 808-822). Springer Berlin Heidelberg.

[16] Bastien, F., Lamblin, P., Pascanu, R., Bergstra, J., Goodfellow, I., Bergeron, A., ... \& Bengio, Y. (2012). Theano: new features and speed improvements. arXiv preprint arXiv:1211.5590.

[17] Goodfellow, I. J., Warde-Farley, D., Lamblin, P., Dumoulin, V., Mirza, M., Pascanu, R., ... \& Bengio, Y. (2013). Pylearn2: a machine learning research library. arXiv preprint arXiv:1308.4214.

[18] Srivastava, N., Hinton, G., Krizhevsky, A., Sutskever, I., \& Salakhutdinov, R. (2014). Dropout: A simple way to prevent neural networks from overfitting. The Journal of Machine Learning Research, 15(1), 1929-1958.

\end{document}